\documentclass[conference]{IEEEtran}
\IEEEoverridecommandlockouts
\usepackage{cite}
\usepackage{url}
\usepackage{amsmath,amssymb,amsfonts}
\newcommand\myfrac[2]{\frac{\displaystyle #1}{\displaystyle #2}}
\usepackage{algorithm}
\usepackage{algorithmicx}
\usepackage{algpseudocode}
\usepackage{graphicx}
\graphicspath{ {./} }
\usepackage{textcomp}
\usepackage{xcolor}

\usepackage{tikz}
\usetikzlibrary{matrix,chains,positioning,decorations.pathreplacing,arrows}
\usetikzlibrary{decorations.text}
\usetikzlibrary{decorations.pathmorphing}
\usetikzlibrary{fit, arrows.meta}
\usepackage{pgfplots}
\pgfplotsset{compat=1.14}

\tikzset{%
	every neuron/.style={
		circle,
		draw,
		minimum size=1cm
	},
	neuron missing/.style={
		draw=none, 
		scale=4,
		text height=0.333cm,
		execute at begin node=\color{black}$\vdots$
	},
	every signode/.style={
	circle,
	fill,
	draw,
	minimum size=0.1cm
	},
}

\def\BibTeX{{\rm B\kern-.05em{\sc i\kern-.025em b}\kern-.08em
    T\kern-.1667em\lower.7ex\hbox{E}\kern-.125emX}}
\begin{document}

\title{Evolving Spiking Neural Networks for Nonlinear Control Problems
}

\author{\IEEEauthorblockN{Huanneng Qiu}
\IEEEauthorblockA{\textit{School of Engineering and Information Technology} \\
\textit{The University of New South Wales at} \\
\textit{the Australian Defence Force Academy} \\
Canberra, Australia \\
huanneng.qiu@student.unsw.edu.au}
\and
\IEEEauthorblockN{Matthew Garratt}
\IEEEauthorblockA{\textit{School of Engineering and Information Technology} \\
\textit{The University of New South Wales at} \\
\textit{the Australian Defence Force Academy} \\
Canberra, Australia \\
m.garratt@adfa.edu.au}
\and
\IEEEauthorblockN{David Howard}
\IEEEauthorblockA{\textit{Robotics and Autonomous Systems Group} \\
\textit{DATA61 at CSIRO} \\
Brisbane, Australia \\
david.howard@data61.csiro.au}
\and
\IEEEauthorblockN{Sreenatha Anavatti}
\IEEEauthorblockA{\textit{School of Engineering and Information Technology} \\
\textit{The University of New South Wales at} \\
\textit{the Australian Defence Force Academy} \\
Canberra, Australia \\
s.anavatti@adfa.edu.au}

}

\maketitle

\begin{abstract}

Spiking Neural Networks are powerful computational modelling tools that have attracted much interest because of the bioinspired modelling of synaptic interactions between neurons.
Most of the research employing spiking neurons has been non-behavioural and discontinuous.  Comparatively, this paper presents a recurrent spiking controller that is capable of solving nonlinear control problems in continuous domains using a popular topology evolution algorithm as the learning mechanism. 
We propose two mechanisms necessary to the decoding of continuous signals from discrete spike transmission: 
(i) a background current component to maintain frequency sufficiency for spike rate decoding, and (ii) a general network structure that derives strength from topology evolution. 
We demonstrate that the proposed spiking controller can learn significantly faster to discover functional solutions than sigmoidal neural networks in solving a classic nonlinear control problem.

\end{abstract}


\section{Introduction} \label{intro}

Artificial Neural Networks (ANNs) are learning based computational systems inspired by the structure of animal brains. 
The current generation of neural networks which are dominant in the computational intelligence community are sigmoidal neural networks, 
whose neurons consist of two components: a weighted sum of inputs and a sigmoidal activation function generating the output accordingly. 
Despite their exceptional performance in a variety of tasks, they are vaguely related to their biological counterparts and do not approximate the signal transmission in biological neural systems.

The idea of Spiking Neural Networks (SNNs) \cite{Maass1997}, therefore, is to bridge the gap between neuroscience and computational intelligence, by employing biologically realistic neuron models to carry out computation. However, over the past decade most of the work analysing information processing in SNNs has been focused on non-behavioural functionality, e.g., character recognition \cite{Rice2009} and approximation \cite{Abbott2016}. 
Meanwhile, there is also a rising interest in behaviourally functional SNNs, which addresses neural activities in closed-loop interaction with the environment \cite{Floreano2003, Batllori2011, Howard2014}.

The most apparent advantage of \emph{neurocontrollers} is that neural networks can learn to perform satisfactory tasks without explicit models of the plants. This is highly preferred in situations where accurate models are difficult to obtain. 
SNNs can be more suitable controllers than non-spiking neural systems because: (i) they are bioinspired plastic learning architectures and can provide faster information processing as observed in biological neural systems \cite{Paugam-Moisy2012}; (ii) they are fundamentally computationally more powerful than non-spiking neural systems \cite{Maass1997};
and (iii) massive network implementation on hardware has already been realised \cite{Merolla2014, Seo2011}, as spiking neuron models can be represented using simple electric circuits.

One issue of SNN implementation is that the \emph{learning} process is challenging using gradient based methods, partly due to the difficulty of extracting gradient information from discrete events, but also because of the recurrent architecture that is essential to preserve memory through internal connections.  
In this paper, we use the popular NEAT algorithm \cite{Stanley2002} as the learning mechanism to generate action selection policies and consequently to seek functional network compositions. NEAT is an ideal neuroevolution strategy due to (i) the efficacy of automatic topology altering along with connection weight; and (ii) compatibility with the discrete nature of SNNs.

The main purpose of this work is to derive continuous motion actions from discrete spike trains in a reinforcement learning task. We claim this has not yet been fully resolved because of two problems: decrease of spike activities through synapses and low resolution of conventional decoding methods. 
Therefore, we construct a recurrent spiking network with two essential settings: a background current to deal with frequency loss, and  a decoding method based on a weighted firing rate. We show that the proposed spiking controller has better performance than its sigmoidal counterpart in solving the classic pole balancing problem.

Organization of the rest of this paper is as follows. Section \ref{snn}  will briefly describe the spiking neuron model and the mechanism to preserve spike activities, followed by a review of the NEAT algorithm in Section \ref{neat}. Section \ref{experiment} presents the experimental setup to test the controller learning algorithm and results of both spiking and non-spiking approaches. Finally, Discussion and Conclusion will be covered in Section \ref{discuss} and Section \ref{conclusion}.

\section{Computation with Spikes} \label{snn}

Unlike traditional neural networks, the information transmission in SNNs relies on electrical pulses. These so-called \emph{spikes} are discrete events that occur at certain points in time. The information carried by spikes is not by the amplitude or form, but rather by the number and the timing of spikes.

A spike is fired when the \emph{membrane potential} of the \emph{presynaptic} neuron exceeds its threshold; then it will travel through the synapses and arrive at all forward-connected \emph{postsynaptic} neurons. One incoming spike would increase the membrane potential of the neuron, which will decrease gradually until it reaches the resting potential if no other arriving spike is observed within a certain period. Therefore, a cluster of incoming spikes would be necessary for a neuron to generate a spike.

\subsection{Neural Model}

\begin{figure}[htbp]
	\centering
	\includegraphics[width=0.47\textwidth]{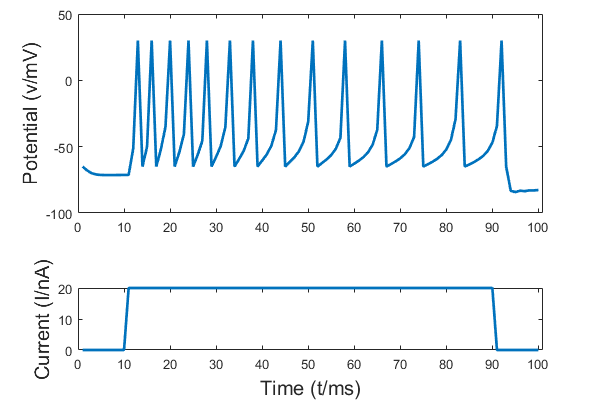}
	\caption{Voltage response of an Izhikevich neuron to a square wave current signal $I$ corresponds to $a = 0.02; b = 0.2; c = -65; d = 2$ and $v_t = 30$ mV.}
	\label{izhi}
\end{figure}

Several spiking neuron models have been proposed over past decades \cite{Gerstner2002}. 
The two-dimensional Izhikevich model \cite{Izhikevich2003} is used in this paper, because of its simplicity whilst being able to produce rich firing patterns to simulate biological neurons by adjusting only a few parameters.
This model is formulated by two ordinary differential equations:

\begin{align}
\dot{v} &= 0.04v^2 + 5v +140 -u + I \\
\dot{u} &= a(bv - u)
\end{align}
with after-spike resetting following:
\begin{equation}
\text{if } v \geq v_t \text{, then}
\left \{
	\begin{array}{l}
		v =c \\
		u = u + d
	\end{array}
\right.
\end{equation}
where $v$ represents the membrane potential of the neuron; $u$ represents a recovery variable; $I$ represents the synaptic current injected into the neuron; $v_t$ is a threshold value; and $a, b, c$ and $d$ are dimensionless parameters to form different spike patterns \cite{Izhikevich2003}. Fig. \ref{izhi} shows the voltage response of an Izhikevich neuron when injected with a square wave current signal.

Firing time $t^{(f)}$ is defined as the moment of $v$ crossing threshold $v_t$ from below:
\begin{equation}
t^{(f)}: v(t^{(f)}) = v_t \text{ and } \dot{v}(t^{(f)}) > 0
\end{equation}
A spike train is then denoted as the sequence of spike times: 
\begin{equation}
s(t) = \sum_f \delta(t-t^{(f)})
\end{equation}
where $\delta(t)$ is the Dirac $\delta$ function\footnote{\url{https://en.wikipedia.org/wiki/Dirac_delta_function}}. 

\subsection{Spike Transmission}

When interfacing the spiking controller with its plant, one critical problem we should consider is how to decode spikes into output commands and encode sensing data into spikes. 

The information representation of spiking neural systems can be distinguished between \emph{rate coding} and \emph{temporal coding} schemes \cite{Theunissen1995}. In a rate coding scheme, neural information is encoded into the number of spikes occurring during a given time window, aka. firing \emph{rate} of spikes. In a temporal coding scheme, the context is encoded into the exact \emph{timing} between presynaptic and postsynaptic spikes. 

In this paper the rate coding scheme is adopted, as it is easier to implement. Mean firing rate is averaged during a given time window, which will be further decoded to generate the output signal.

Encoding of sensing data will go through two stages. 
Input variables are first normalized within the range of [0,1], after which, the rescaled signal will be linearly converted into a current value that will be injected into forward-connected neurons through synapses. 
This current encoding method is commonly used in literature, and is compatible with different coding schemes, as the neuron receiving larger input currents will not only fire at higher rate, but also be earlier to spike \cite{Gamez2012}. 

\begin{figure}[htbp]
	\centering
	\includegraphics[width=0.47\textwidth]{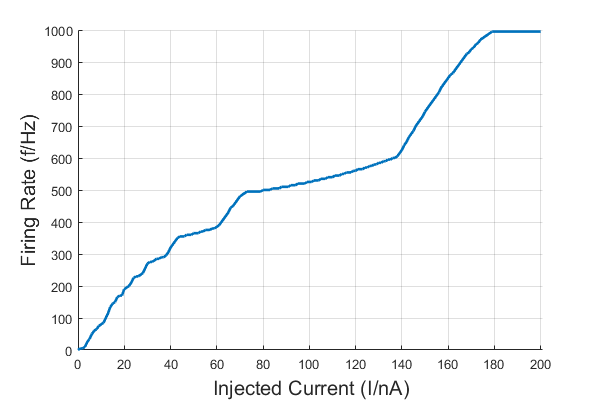}
	\caption{$f$-$I$ curve of an Izhikevich neuron with parameters as: $a = 0.02; b = 0.2; c = -50; d = 2$ and $v_t = 30$ mV.}
	\label{fi_curve}
\end{figure}

In a rate coding framework, the neuron will fire at a steady-state firing rate $f$, given the injected current $I$ is fixed in time. Therefore, the spike train frequency can be defined as a function of the magnitude of the input stimulus. 
Fig.~\ref{fi_curve} shows the curve of mean firing rate of an Izhikevich neuron when the input current varies from zero to 200 nA.

It is critical to maintain sufficient spike transmission in order to calculate a smooth firing rate. However, as mentioned, a cluster of incoming spikes are generally required to force a neuron to spike. Therefore, there tends to be a transmission loss as spikes travel through the synapses. In this paper, we try to resolve this problem by simply adding a background current, with which neurons will fire at a certain frequency even if there is no other input stimulus. Such an effect can also be found in biology, where synaptic background noise can significantly affect the neuron characteristics \cite{Fellous2003}.

\section{Neuroevolution} \label{neat}

Despite the success of gradient methods in training traditional multilayer neural networks, their implementation on SNNs is still problematic when extracting gradient information from output spike times. 
Instead, population-based neuroevolution(NE) is an ideal learning architecture for our settings -- recurrent network learning and evolution of network topology.

\subsection{Evolutionary Algorithms (EAs)}

EAs \cite{Eiben2015} are an abstraction of the natural evolution process. A general scheme of EAs is given in Algorithm~\ref{evoAlg}. One of the cornerstones of EAs is competition based selection, which is widely known as \emph{survival of the fittest} -- individuals that are more adapted to the environment have a higher chance to create offspring or survive. Another basis lies in phenotype variation. To generate different individuals, variation operations (i.e. \emph{recombination} (also termed \emph{crossover}) and \emph{mutation}) are applied, producing new individuals that loosely resemble their parents. The combination of variation and selection thereby leads to a population that is better adapted to the environment or better able to complete a given task.

\begin{algorithm}
	\caption{General Scheme of Evolutionary Algorithms}
	\label{evoAlg}
	\begin{algorithmic}[1]
		\State INITIALISE a population of candidate solutions
		\State EVALUATE each individual \& ASSIGN fitness values
		\While {(terminate condition not met)}
		\State SELECT parents
		\State RECOMBINATION parents
		\State MUTATE offspring
		\State EVALUATE new candidates
		\State SELECT individuals for the next generation
		\EndWhile
	\end{algorithmic}
\end{algorithm}

However, EAs are likely to converge around one single solution as evolution goes on -- a phenomenon known as \emph{genetic drift} \cite{Eiben2015}, which can lead to premature convergence and consequently getting stuck at local optimum, because the population may quickly converge on whatever network that happens to perform best in the initial population. 
Therefore, the NEAT algorithm \cite{Stanley2002} is introduced in this paper to overcome this problem.

\subsection{NeuroEvolution of Augmenting Topology (NEAT)}

NEAT is a powerful evolutionary approach for neural network learning which evolves network topologies along with connection weights. The efficacy of NEAT is guaranteed by: (i) historical markings to solve the variable-length genome problem; (ii) speciation to protect innovation and preserve network diversity, to avoid premature convergence; and (iii) incremental structural growth to avoid troublesome hand design of network topology.

Typical EAs are difficult to genetically crossover neural networks with variant topologies, because they can only operate within fixed-sized genome space.
Recombination of divergent genomes tends to produce damaged offspring. 
However, similar network solutions sharing similar functionalities can be encoded using completely different genomes -- a phenomenon known as the \emph{Competing Convention Problem} \cite{Stanley2002}.
To address this problem, NEAT uses historical markings which act as artificial evidence to track the origin of genes. When two genes share the same historical marking, they are categorised as alleles. 
Therefore, NEAT can match up genomes representing similar network structures and allow mating in a rationale manner.

NEAT also uses an explicit fitness sharing scheme \cite{Eiben2015} as a population management approach to preserve network multi-modality. Historical markings are used as a measurement of the genetic similarity of network topologies, based on which, genomes are speciated into different \emph{niches} (also termed \emph{species}). 
Individuals clustered into the same species will share their fitness score together \cite{Eiben2015}. 
The fitness of each individual is scaled according to the number of individuals falling in the niche. As competition within the species becomes more intense, solutions will have a lower fitness. Thus, the sparsely populated species will become more attractive, avoiding any one taking over.
Therefore, innovations will be protected within niches to have time to optimize.

Finally, an incremental growth mechanism is used in NEAT to discover the least complex effective neural topology, by beginning searching minimal network structure and gradually expanding to more complex networks during evolution.

The NEAT algorithm has been successfully applied to different control problems \cite{Stanley2002, Pardoe2005, Shepherd2010}. However, implementation of NEAT has been restricted to traditional neural systems, with one exception \cite{Vandesompele2016} targeting spiking networks.

\subsection{Speciation Measurement}
NEAT use a compatibility distance function $\delta$ to determine the similarity of network solutions. When the distance between any two individuals is smaller than a threshold $\delta_t$, they are categorized into the same species. The compatibility distance $\delta$ is defined as:
\begin{equation}
\delta = \frac{c_1 E}{N} + \frac{c_2 D}{N} + c_3 \overline{W} \label{compat}
\end{equation}
where $E$ and $D$ denote the number of \emph{excess} and \emph{disjoint} genes; $\overline{W}$ denotes the average connection weight difference; $N$ is the total number of genes; $c_1$, $c_2$ and $c_3$ are user-defined coefficients for altering the significance of these factors.

In our second task illustrated later, we will apply a slight modification to this function by taking account of the influence of the decoding method. There will be a more detailed description in Section \ref{pbnv}.

\section{Experiments} \label{experiment}
We evaluate the system's performance using a classic nonlinear control benchmark -- the pole balancing problem (also known as the inverted pendulum problem). 
This problem is not only inherently unstable, but also capable of varying degrees of complexity by limiting the state variables provided to the controller, which makes it ideal for designing and testing nonlinear control methods.

Previous attempts to solve this problem using SNNs with fixed-topology can be found in \cite{Vasu2017a, Kang2017}. In this paper, we take a different approach using NEAT. 
Results of the proposed spiking controller are benchmarked against the original sigmoidal counterpart. 
The original NEAT C++ source code is available publicly\footnote{\url{http://nn.cs.utexas.edu/?neat-c}}, which is tailored to be amenable to our network model. All the experiments are programmed in C++ and performance analysis is carried out using MATLAB.

\subsection{Benchmark Problem}

\begin{figure}[htbp]
	\centering
	\includegraphics[width=0.43\textwidth]{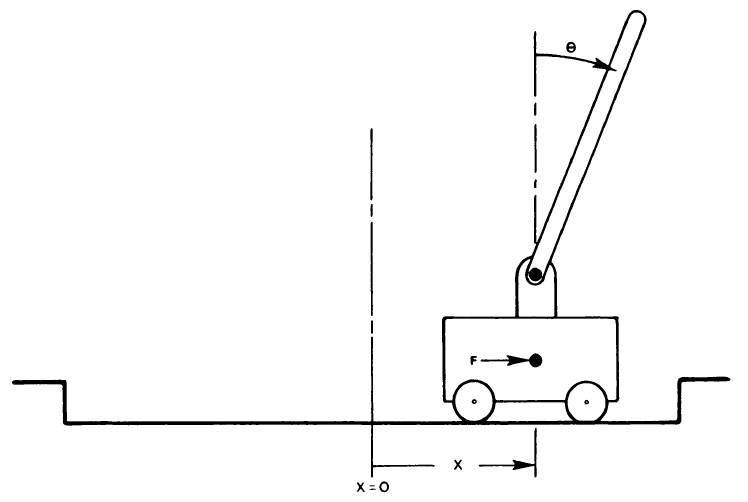}
	\caption{Cart-pole system, taken from \cite{Barto1983}.}
	\label{pole}
\end{figure}

Fig. \ref{pole} shows the cart-pole system to be controlled, which consists of a cart that can move left or right within a bounded one-dimensional track, and a pole that is hinged to the cart.
The problem is to balance the pole upright for as long as possible by applying a force $F_t$ to the cart parallel to the track. 
The system has four state variables: 
\begin{itemize}
	\item[] $x$ -- cart position
	\item[] $\theta$ -- pole angle
	\item[] $\dot{x}$ -- cart velocity
	\item[] $\dot{\theta}$ -- pole angular velocity
\end{itemize}

For simplicity, we neglect the friction of cart on track and that of pole on cart. 
The system is then formulated by two nonlinear differential equations \cite{Barto1983}:

\begin{align}
\ddot{\theta}_t	&= \myfrac{g \sin \theta + \cos \theta (\frac{-F_t - m_p l \dot{\theta}^2 \sin \theta}{m_c + m_p})}
{l (\frac{4}{3} - \frac{m_p \cos^2 \theta}{m_c + m_p})} \label{theta_ddot}\\
\ddot{x}_t &= \frac{F_t + m_p l (\dot{\theta}^2 \sin \theta - \ddot{\theta} \cos \theta)}{m_c + m_p} \label{x_ddot}
\end{align}
where $g = 9.8$ m/s\textsuperscript{2} denotes the gravitational acceleration; $m_c = 1.0$ kg denotes the mass of cart and $m_p = 0.1$ kg denotes the mass of pole; $l = 0.5$ m is half length of the pole. 

The discrete form of state variables are updated following:

\begin{equation} \label{x_n_theta}
\left \{
\begin{array}{lcl}
	x_{t+1} & = & x_t + \tau \dot{x}_t \\
	\dot{x}_{t+1} & = & \dot{x}_t + \tau \ddot{x}_t \\
	\theta_{t+1} & = & \theta_t + \tau \dot{\theta}_t \\
	\dot{\theta}_{t+1} & = & \dot{\theta}_t + \tau \ddot{\theta}_t 
\end{array}
\right.
\end{equation}
with $\tau$ representing the time step.

A force generated by the spiking controller at each time step will be used to update the state variables following \eqref{theta_ddot}, \eqref{x_ddot} and \eqref{x_n_theta}. 
A failure signal is generated when the cart reaches the track boundary, which is $\pm 2.4$ meters from the track centre, or if the pole tilts beyond the failure angle, which is $\pm 12$ degrees (or about 0.21 radian) from the vertical.

\subsection{Experimental Setup}

We first start with the basic balancing task with complete state variables. This Markovian problem can act as a base performance measurement before we go to the more challenging non-Markovian version without velocity information. In both tasks the cart-pole system model is unknown to the spiking controller. 

The controller contains a population of 150 networks, which will be evolved using a combined EA, including a (150, 150) Evolution Strategy \cite{Eiben2015} plus species-based Elitism. Per epoch, the champion of each species is duplicated when the number of networks in that species is larger than 5. The best $20\%$ of networks in each species are allowed to reproduce, after which, all parents are discarded and the remaining 150 offspring will form the next generation.

Each network will be evaluated and assigned a fitness value. Fitness is defined as the number of time steps that the balanced criteria is not violated. Otherwise a failure signal is generated and evaluation is moved on to the next network.

\subsection{Pole Balancing with Velocity}

The first task is to balance the pole with velocity information. At initialization, each network consists of 5 spiking neurons -- 4 input nodes each receiving one state variable and one output node generating a force applied to the cart; and 4 connections each connecting one input node to the output node, respectively. 
We use a probabilistic rate coding method here. The normalized input variables are encoded as a probability to generate a spike. At each time step, this probability will be used to determine the firing status of the input neuron.
Firing rate is then calculated based on the output node. A binary force ($F_t = \pm 10$ Newtons) is generated each time to be applied to the cart.

During evolution, hidden nodes are allowed to be added with a probability of 0.03. Connections are added with a probability of 0.1. 
The time step $\tau$ in \eqref{x_n_theta} is set to 0.02 seconds. 
A successful solution is identified when it is able to balance the pole for 100,000 time steps, which is equivalent to around 30 minutes of simulated time.

We apply NEAT to evolve both SNNs and sigmoidal networks. 
Each test is run for 60 episodes. Table \ref{tabsa} shows a summary of fewest generations needed to complete the task. A failure run means the controller fails to find a solution within 100 generations.
Both approaches failed once in 60 runs. 
For those successful runs, it is interesting to notice that the spiking controller takes fewer generations to solve the task. 

\begin{figure}[!htbp]
	\centering
	\includegraphics[width=0.48\textwidth]{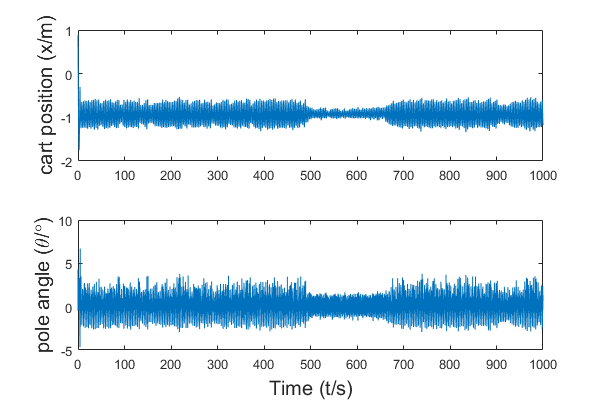}
	\caption{NEAT original: cart position $x$ and pole angle $\theta$ of the pole balancing task with velocity information.}
	\label{run_ann}
\end{figure}

\begin{figure}[!htbp]
	\centering
	\includegraphics[width=0.48\textwidth]{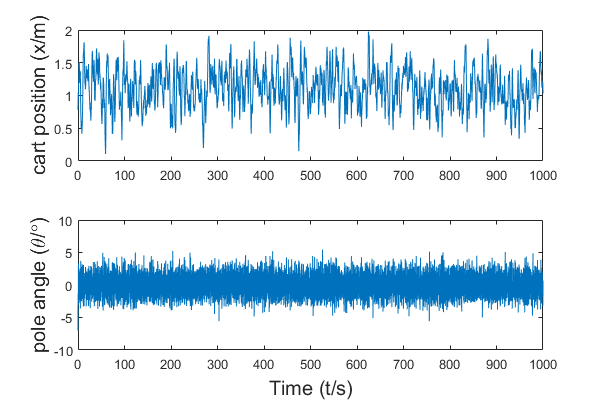}
	\caption{NEAT-SNN: cart position $x$ and pole angle $\theta$ of the pole balancing task with velocity information.}
	\label{run_snn}
\end{figure}

Example runs of both tests are shown in Fig.~\ref{run_ann} and Fig.~\ref{run_snn}.
The spiking approach has more jerky outputs due to the probability coding method we use to encode sensing data.
Smaller output errors can be observed when the magnitude of the force $F_t$ is set smaller.

\subsection{Pole Balancing without Velocity} \label{pbnv}

Our main task is to balance the pole without velocity information. Instead of using bang-bang control, we use a continuous output force within [-10, 10] Newtons for this more challenging problem. A normalized force $F_n$ is first calculated based on a weighted sum of firing rates from connected neurons following a modified sigmoid function, which is then scaled and shifted to generate the force $F_t$:
\begin{align}
F_n &= \frac{1}{1 + \exp (-\sigma \sum w_i r_i)} \label{fn} \\
F_t &= 10 (2 F_n - 1) \label{ft}
\end{align}
where $\sigma$ is a positive decay variable, which will be automatically tuned during evolution; $r_i$ denotes the firing rate of the \textit{i\textsuperscript{th}} connected neuron and $w_i$ denotes the corresponding connection weight. 

\begin{table}[tbp]
	\caption{Fewest Generations Required to Find A Successful Solution for Markovian Pole Balancing Problem}
	\begin{center}
		\begin{tabular}{|c|c|c|c|c|c|}
			\hline
					& Best & Worst & Median$^{\mathrm{a}}$ & Mean$^{\mathrm{a}}$ & Failure Rate \\
			\hline 
			NEAT original & 3 & 55 & 17.5 & 21.68 & 1/60 \\
			\hline
			NEAT-SNN & 1 & 92 & 9.5  & 16.55 & 1/60 \\
			\hline
			\multicolumn{6}{l}{$^{\mathrm{a}}$Values are calculated assuming failure runs take 101 generations.}
		\end{tabular}
		\label{tabsa}
	\end{center}
\end{table}

The aforementioned compatibility distance function \eqref{compat} is modified by considering $\sigma$ in \eqref{fn}. A component of $\sigma$ difference (denoted as $D_\sigma$) is added to the original compatibility distance:
\begin{equation}
\delta = \frac{c_1 E}{N} + \frac{c_2 D}{N} + c_3 \overline{W} + c_4 D_\sigma
\end{equation}

\begin{figure}[htbp]
	\centering
	\begin{tikzpicture}[x=1.5cm, y=1.5cm, >=stealth]
	
	\foreach \m/\l [count=\y] in {1,2}
	\node [every signode/.try, neuron \m/.try] (input-\m) at (0,0.5+\y) {};
	
	\foreach \m [count=\y] in {1,2,3}
	\node [every neuron/.try, neuron \m/.try ] (hidden-\m) at (2,\y) {};
	
	\foreach \m [count=\y] in {1}
	\node [every signode/.try, signode \m/.try ] (output-\m) at (4,1+\y) {};
	
	\draw [<-] (input-1) -- ++(-1,0)
	node [above, midway] {$\theta_t$};
	
	\draw [<-] (input-2) -- ++(-1,0)
	node [above, midway] {$x_t$};
	
	\draw [->] (output-1) -- ++(1,0)
	node [above, midway] {$F_t$};
	
	\foreach \i in {1,...,2}
	\foreach \j in {1,...,2}
	\draw [->] (input-\i) -- (hidden-\j);
	\draw [->] (input-2) -- (hidden-3);
	
	\draw [->] (hidden-2) -- (output-1) 
	node [above, midway, sloped] {$w_2 r_2$};
	\draw [->] (hidden-3) -- (output-1) 
	node [above, midway, sloped] {$w_1 r_1$};
	
	\draw[->,shorten >=1pt] (hidden-2) to [out=270,in=315,loop,looseness=5] (hidden-2);
	\draw[->,shorten >=1pt] (hidden-3) to [out=45,in=90,loop,looseness=5] (hidden-3);
	
	\draw[->,shorten >=1pt] (hidden-1) to [out=120,in=-120,loop,looseness=1] (hidden-2);
	\draw[->,shorten >=1pt] (hidden-1) to [out=0,in=0,loop,looseness=0.8] (hidden-3);

	\end{tikzpicture}
	\caption{Spiking network topology that allows internal recurrence. Network inputs consist of cart position $x_t$ and pole angle $\theta_t$. Output force $F_t$ is calculated based on a weighted sum of firing rates $\sum w_i r_i$, in which $r_i$ denotes the firing rate of the \textit{i\textsuperscript{th}} connected neuron and $w_i$ denotes the corresponding connection weight.}
	\label{snn_topo}
\end{figure}
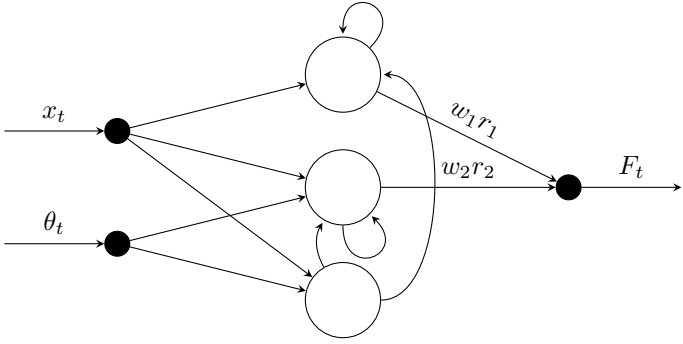

Fig.~\ref{snn_topo} shows a possible network topology during evolution. Inputs are cart position and pole angle. Connections and nodes will be added based on a probability of 0.1 and 0.03. Recurrence is allowed within spiking neurons, facilitating internal calculation of derivatives. 

The initial state variables are set to 0, except the pole angle is set to 3 degrees.
Time step $\tau$ is set to 0.01 seconds. Successful solutions are dictated if the pole is balanced for 5000 time steps. 

Similarly, we apply NEAT to the spiking controller and compare the results against the original NEAT. 
A summary is shown in Table \ref{tabsa_non}. 
The proposed spiking controller is essentially better than sigmoidal networks in solving this problem. It requires fewer generations to find a functional solution. It also has a lower failure rate over 60 runs, showing the potential to be more adaptive. 
Further, the Mann-Whitney \emph{U}-test is used to assess the statistic difference between the two sets of samples. The $p$ value is smaller than 0.01, showing that the spiking controller has significantly better performance. 

\begin{figure}[!bp]
	\centering
	\includegraphics[width=0.45\textwidth]{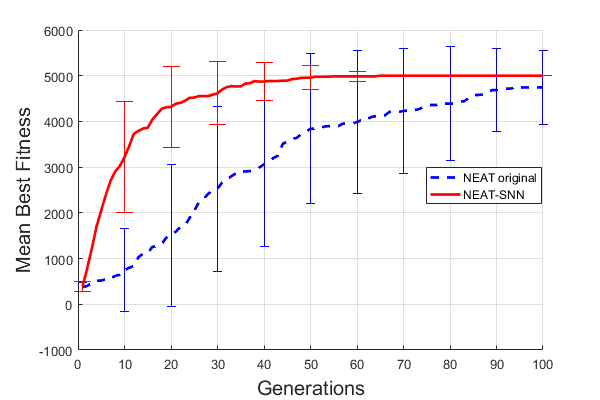}
	\caption{Best networks' mean fitness values in progress over 60 runs, with error bars denoting the standard deviation.}
	\label{best_fit}
\end{figure}

To visualize the evolution progress, we average the best networks' fitness values over 60 runs. Fig.~\ref{best_fit} shows the mean and standard deviation of fitness values of both tests at successive generations. In the beginning, only some of the networks can find a path to optimization, thus introducing a large fitness deviation. As evolution goes on, individuals with higher fitness values will gradually take over the entire population.

\begin{table}[tbp]
	\caption{Fewest Generations Required to Find A Successful Solution for Non-Markovian Pole Balancing Problem}
	\begin{center}
		\begin{tabular}{|c|c|c|c|c|c|}
			\hline
			& Best & Worst & Median$^{\mathrm{a}}$ & Mean$^{\mathrm{a}}$ & Failure Rate \\
			\hline 
			NEAT original & 8 & 96 & 44 & 50.35 & 7/60 \\
			\hline
			NEAT-SNN & 5 & 65 & 21.5  & 24.70 & 0/60 \\
			\hline
			\multicolumn{6}{l}{$^{\mathrm{a}}$Values are calculated assuming failure runs take 101 generations.}
		\end{tabular}
		\label{tabsa_non}
	\end{center}
\end{table}

\begin{figure}[htbp]
	\centering
	\includegraphics[width=0.47\textwidth]{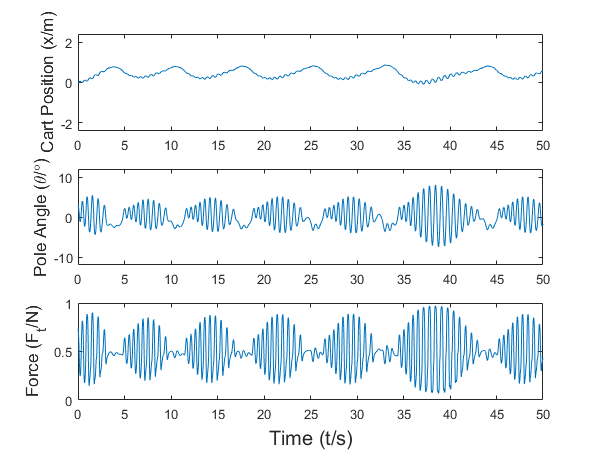}
	\caption{NEAT original: cart position $x$, pole angle $\theta$ and force $F_t$ applied to the cart of the non-Markovian pole balancing task (without velocity information).}
	\label{run_ann_non}
\end{figure}

\begin{figure}[htbp]
	\centering
	\includegraphics[width=0.47\textwidth]{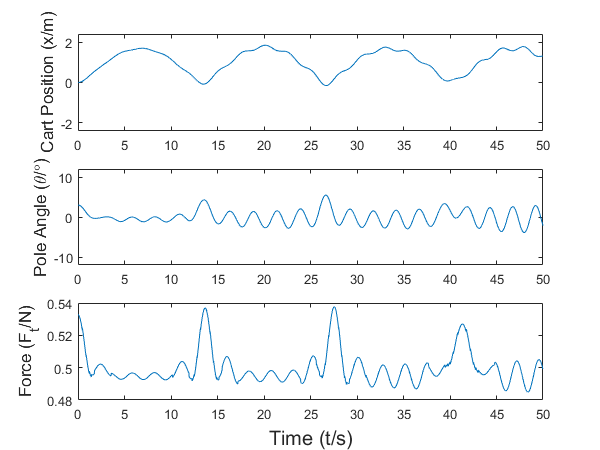}
	\caption{NEAT-SNN: cart position $x$, pole angle $\theta$ and force $F_t$ applied to the cart of the non-Markovian pole balancing task (without velocity information).}
	\label{run_snn_non}
\end{figure}

Fig.~\ref{run_ann_non} and Fig.~\ref{run_snn_non} show the status of cart position, pole angle and the output force applied to the cart during a successful run. In both approaches, all these three variables are oscillating around a certain point.

\section{Discussion} \label{discuss}

The design of functional SNNs is considered to be difficult, because SNNs behave as complex systems with transient dynamics \cite{Paugam-Moisy2012}. Therefore, parameter setting of spiking network models to solve a given task is non-trivial and still not entirely resolved. Apart from the neuron model, synaptic dynamics with transmission delay also acts as a significant component to the computation power of SNNs.
We argue evolution can be beneficial to the automatic tuning of these network parameters.  

Another active implementation problem is to develop a meaningful approach to transform continuous variables into a spike representation, and vice versa.
Rate coding methods either require a relatively long time or large group of homogeneous neurons to calculate a smooth firing rate.  
On the other hand, temporal coding is considered to be more biologically plausible. 
Although not yet fully understood, this coding scheme is evidentially able to provide faster signal processing and hence less reaction time.

\section{Conclusion} \label{conclusion}
NEAT is a performance-guaranteed neuroevolution method. This algorithm is tailored in this work to be amenable to SNNs. 
Through the experiments, we have demonstrated that SNNs can solve continuous control problems by maintaining sufficient spike activities and decoding from weighted spike frequencies. 
The proposed spiking controller has shown to be more rapid than sigmoidal networks in finding functional solutions. 
Our work is a first step toward robot control and the results have encouraged further experimentation on more challenging dynamic systems. 
We expect ensemble-based implementations \cite{Howard2016a} would provide redundancy to neuroevolution and hence achieve robust performance.
We are also looking at integration of synaptic plasticity as a means to online-offline hybrid learning \cite{Howard2012a}.
Our end target is to implement the spiking controller on flight systems such as flapping wing Micro Air Vehicles.

\end{document}